\documentclass[journal,twoside,web]{ieeecolor}
\usepackage{generic}

\usepackage[utf8]{inputenc}
\usepackage[T1]{fontenc}
\usepackage{amsmath,amssymb,amsfonts}
\usepackage{mathrsfs}
\usepackage{algorithm}
\usepackage{algpseudocode}
\usepackage{graphicx}
\usepackage{float}
\usepackage{caption}
\usepackage{subcaption}
\usepackage{wrapfig}
\usepackage{tikz}
\usepackage{listings}

\usepackage[table]{xcolor}
\usepackage{booktabs}
\usepackage{multirow}
\usepackage{colortbl}
\usepackage{mdframed}
\usepackage{siunitx}

\usepackage[colorlinks,urlcolor=blue,linkcolor=blue,citecolor=blue]{hyperref}

\usepackage{fontawesome5}

\graphicspath{{./figures/}}

\definecolor{lightgray}{RGB}{248,248,248}
\definecolor{ash}{RGB}{200,200,200}
\definecolor{skyblue}{RGB}{245,250,255}
\definecolor{brown}{RGB}{139,69,19}
\definecolor{headerblue}{RGB}{220,230,250}
\definecolor{systemblue}{RGB}{70,130,200}
\definecolor{usergreen}{RGB}{34,139,34}
\definecolor{outputgray}{RGB}{95,95,95}
\definecolor{lightblue}{RGB}{240,248,255}
\definecolor{lightgreen}{RGB}{240,255,240}
\definecolor{prunedorange}{RGB}{255,140,0}
\definecolor{bestpurple}{RGB}{138,43,226}
\definecolor{lightorange}{RGB}{255,248,240}
\definecolor{lightpurple}{RGB}{248,240,255}
\definecolor{imagegray}{RGB}{248,248,250}

\usepackage{tcolorbox}
\tcbuselibrary{skins,breakable}
\tcbset{
  modelbox/.style={
    enhanced,
    colback=skyblue!20,
    colframe=black,
    boxrule=1.2pt,
    arc=12pt,
    outer arc=12pt,
    breakable,
    left=8pt,right=8pt,top=8pt,bottom=8pt,
    width=\textwidth,
    valign=top
  }
}

\def\BibTeX{{\rm B\kern-.05em{\sc i\kern-.025em b}\kern-.08em
    T\kern-.1667em\lower.7ex\hbox{E}\kern-.125emX}}

\begin{document}

\title{GRPO++: Enhancing Dermatological Reasoning under Low Resource Settings}

\author{Ismam Nur Swapnil\textsuperscript{\dag}, 
Aranya Saha\textsuperscript{\dag}, 
Tanvir Ahmed Khan\textsuperscript{\dag},
        Mohammad Ariful Haque
\thanks{\textsuperscript{\dag}Equal Contribution.}%
\thanks{All authors are with the Department of Electrical and Electronic Engineering, Bangladesh University of Engineering and Technology (BUET), Dhaka-1205, Bangladesh.}%
\thanks{Corresponding authors: 
Ismam Nur Swapnil (e-mail: ismamnurswapnil@gmail.com), 
Aranya Saha (e-mail: aranyasaha932@gmail.com), 
Tanvir Ahmed Khan (e-mail: tanvirahmedkhan0601@gmail.com), 
Mohammad Ariful Haque (e-mail: arifulhoque@eee.buet.ac.bd).}
}

\maketitle

\begin{abstract}
Vision-Language Models (VLMs) show promise in medical image analysis, yet their capacity for structured reasoning in complex domains like dermatology is often limited by data scarcity and the high computational cost of advanced training techniques. To address these challenges, we introduce \textbf{DermIQ-VLM}, a VLM developed through a multi-stage, resource-efficient methodology designed to emulate a dermatologist's diagnostic process. Our primary contribution is a modified version of Grouped Relative Policy Optimization (GRPO), called GRPO++, which stabilizes the powerful but data-intensive GRPO framework. Our proposed training pipeline first employs GRPO++ for reasoning-oriented disease recognition, followed by supervised fine-tuning for conversational ability. To mitigate factual errors introduced during this step, we then align the model using Direct Preference Optimization (DPO), leveraging a Knowledge Graph-based system as a scalable proxy for expert preference. A preliminary evaluation on a curated dermatological dataset demonstrates that our proposed methodology yields notable performance gains over standard fine-tuning approaches. These findings validate the potential of our pipeline as a feasible pathway for developing specialized, reliable VLMs in resource-constrained environments.

\end{abstract}


\section{Introduction}

\IEEEPARstart{T}{housands} of skin lesions are assessed annually, driving the global demand for accurate and reliable AI support in dermatology. As clinical workloads rise and diagnoses become more complex, automated systems that not only predict outcomes but also reason and explain are becoming indispensable. Recent Vision–Language Models (VLMs) such as GPT-4o~\cite{hurst2024} and Grok~\cite{xai2025grok3} demonstrate strong multimodal reasoning capabilities in medical Visual Question Answering (VQA)~\cite{lin2021}, paving the way for interpretable diagnostic assistance. Despite this promise, most medical VLMs still rely on shallow, pattern-based explanations that lack transparency, limiting their integration into clinical workflows and diminishing clinician trust~\cite{shojaee2025}.

Supervised Fine-Tuning (SFT) remains the dominant paradigm for adapting foundation models to medical tasks. While SFT achieves strong performance on benchmark datasets, it often causes overfitting and shortcut learning~\cite{ghosh2024}, leading to poor generalization for rare dermatological conditions~\cite{thirunavukarasu2023}. More critically, SFT fails to capture the stepwise diagnostic reasoning process that clinicians naturally follow, making generated responses appear correct but clinically superficial. Reinforcement learning with human feedback (RLHF) has emerged as a solution, offering reasoning support aligned with expert judgment, but it is computationally expensive and depends heavily on high-quality annotations. Chain-of-thought (CoT) fine-tuning~\cite{wei2022} provides interpretable reasoning traces, yet its reliance on costly expert labeling limits scalability. Group Relative Policy Optimization (GRPO)~\cite{shao2024}, an efficient reinforcement learning method that optimizes relative preferences within sampled outputs, has shown promise but remains underexplored in medical VQA. To address these gaps, we propose \textbf{GRPO++}, a scalable variant designed to strengthen reasoning-oriented optimization while remaining practical for clinical deployment.

Another major barrier to adoption lies in hallucinations and factual inaccuracies, which can undermine clinical safety~\cite{kim2025}. To mitigate this, we integrate Knowledge Graph-based Retrieval-Augmented Generation (KG-RAG)~\cite{sanmartin2024}, which grounds responses in a dermatology-specific corpus constructed from reliable medical sources. While retrieval helps reduce unsupported claims, it alone does not guarantee that the model internalizes knowledge for consistent use across contexts. To bridge this gap, we employ Direct Preference Optimization (DPO)~\cite{rafailov2023direct}, which directly aligns the model’s generation preferences with grounded, factually reliable reasoning patterns. By combining KG-RAG with DPO, we achieve both external grounding and internalized reliability, ensuring accurate outputs even in retrieval-free settings.

\begin{figure*}[!t]
    \centering
    \includegraphics[width=\linewidth]{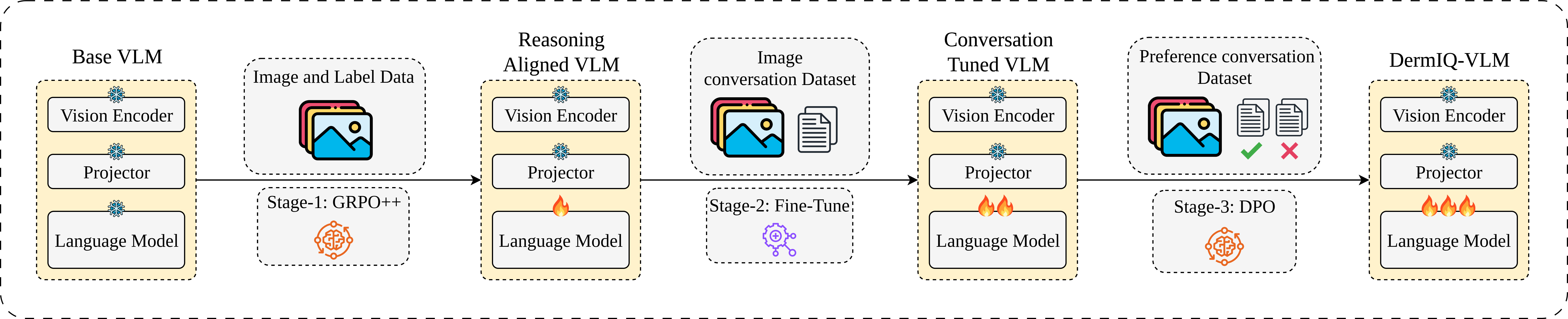}
    \caption{Proposed Training Methodology}
    \label{fig:training_flow}
\end{figure*}

\bigskip

In this work, we introduce \textbf{DermIQ-VLM}, a dermatology-specific VLM designed to deliver interpretable and clinically aligned diagnostic support. Our key contributions are summarized as follows:

\begin{itemize}
\item \textbf{Dataset Curation:} We curate a comprehensive dermatological VQA dataset from trusted clinical sources, systematically structured for training, fine-tuning, and evaluation.
\item \textbf{Reasoning-Oriented Optimization:} We propose \textit{\textbf{GRPO++}}, an enhanced reinforcement learning method that explicitly aligns model reasoning with clinical diagnostic processes, improving interpretability and trustworthiness.
\item \textbf{Low-Resource Training:} We demonstrate that GRPO++ remains effective under constrained response budgets, significantly reducing computational overhead while preserving reasoning quality.
\item \textbf{Clinical Alignment via Grounding:} By integrating \textit{KG-RAG} and \textit{DPO}, we enable the model to both ground responses in medical knowledge and internalize these patterns, resulting in reliable and accurate outputs even without retrieval.
\end{itemize}

\section{Related Works}

Vision–Language Models (VLMs) combine visual and textual inputs for multimodal reasoning in tasks such as image captioning, VQA, and clinical reporting. In healthcare, large language models (LLMs) support question answering, diagnostic reporting, and decision support \cite{kline2022}. Systems like ChatCAD and OphthUS-GPT embed LLMs into diagnostic pipelines, translating visual data into reports \cite{gan2025}. Yet, many efforts remain restricted to single modalities, such as ECGs or chest X-rays \cite{kline2022, wang2024}, underscoring the need for broader multimodal integration.

VLMs have advanced AI in dermatology, radiology \cite{zhang2022}, pathology \cite{chen2023}, and general clinical domains \cite{tu2023, moor2023}. Recent models enable multimodal interpretability, supporting joint reasoning over images and text \cite{liu2023, wang2024qwen2, zhang2024enhanced, pan2025medvlm}. However, supervised fine-tuning (SFT) often produces shallow pattern learning inadequate for complex diagnostics \cite{mccoy2019}, motivating multistage optimization strategies for deeper reasoning.

Interpretability is central for clinical adoption. Chain-of-Thought (CoT) prompting \cite{wei2022} and fine-tuning on structured clinical CoT data \cite{chung2022, jacovi2020} enhance sequential reasoning and coherence. Hallucination remains a barrier \cite{ji2023}, but Retrieval-Augmented Generation (RAG) \cite{lewis2020}, particularly knowledge-graph-based RAG (KG-RAG) \cite{yasunaga2021}, improves factual reliability by grounding outputs in structured knowledge \cite{nicholson2020}.

Alignment methods further refine medical VLMs. Reinforcement Learning from Human Feedback (RLHF) \cite{ouyang2022} has been extended by Direct Preference Optimization (DPO) \cite{rafailov2023direct}, training models to prioritize accurate outputs. Group Relative Policy Optimization (GRPO) \cite{guo2025deepseek}, a variant of Proximal Policy Optimization \cite{schulman2017proximalpolicyoptimizationalgorithms}, has been proposed for structured reasoning, often combined with DPO and KG-RAG for factual alignment.

In dermatology, AI has centered on lesion classification with large datasets \cite{esteva2017, brinker2019, tschandl2018}, achieving high accuracy but limited interactivity. Modern medical VLMs are shifting toward interactive, reasoning-oriented frameworks that deliver interpretable, step-by-step rationales by integrating visual and clinical knowledge for reliable diagnostic support.

\section{Proposed Training Methodology}\label{Proposed Training Methodology}

Our training framework follows a structured process. In stage-1, GRPO++ based reinforcement learning is used to initialize visual disease detection capabilities. In stage-2, Supervised Fine-Tuning (SFT) further enhances multi-turn conversational performance. Finally, in stage-3, we align the model using Direct Preference Optimization (DPO) with the help of Knowledge-Graph (KG-RAG) to improve factual accuracy and reduce hallucinations, using a preference dataset. The complete methodology is summarized in Figure~\ref{fig:training_flow}.

\begin{figure*}[!t]
    \centering
    \includegraphics[width=\textwidth]{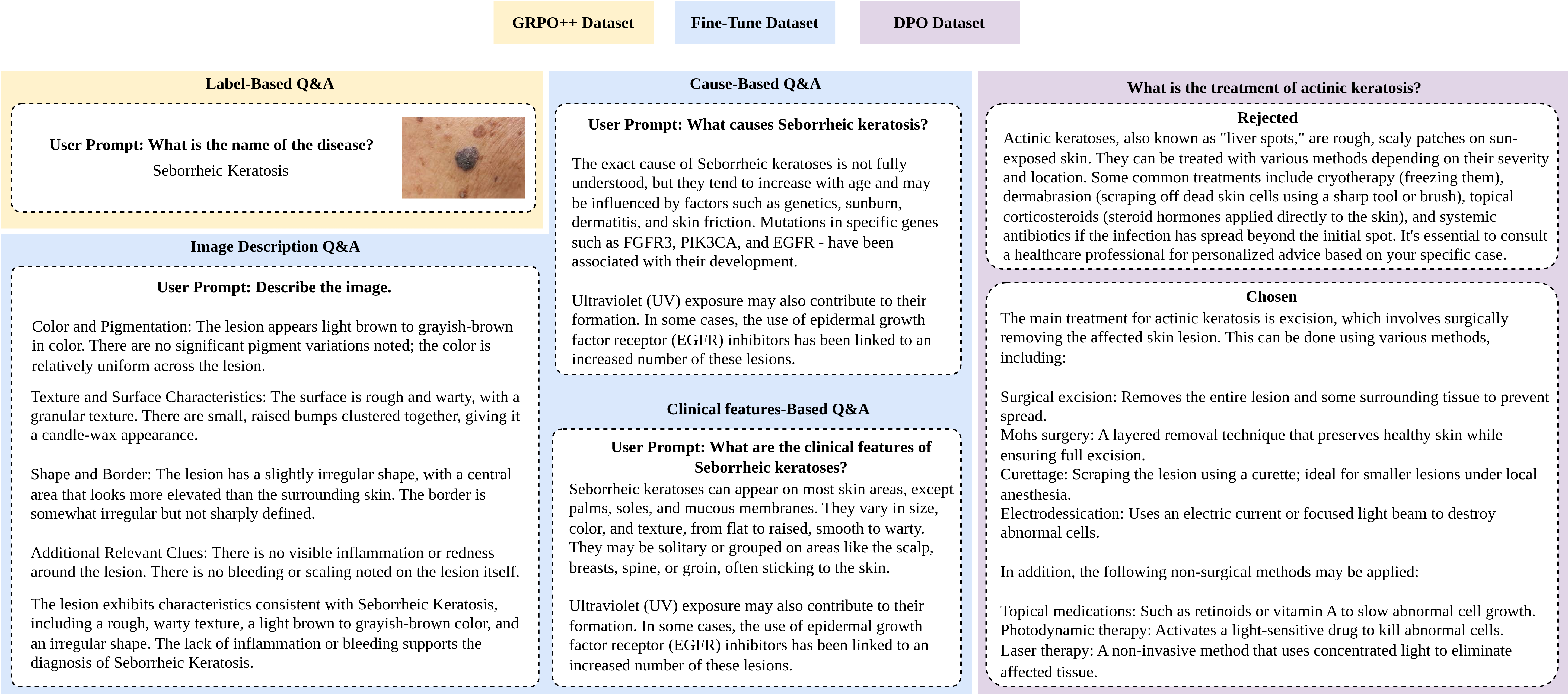}
    \caption{Example Data Points from Our Curated Dataset}
    \label{fig:dataset_example}
\end{figure*}

\subsection{\textbf{Dermatological Dataset Curation for Training}}

\vspace{0.1cm}
\subsubsection{\textbf{Image and Label Dataset for GRPO++}}

We developed a specialized dermatological Visual Question Answering (VQA) dataset to distinguish among seven skin diseases with similar appearances: \textit{Dermatitis}, \textit{Basal Cell Carcinoma}, \textit{Rosacea}, \textit{Psoriasis}, \textit{Actinic Keratosis}, \textit{Seborrheic Keratosis}, and \textit{Melanoma}. Images from the \textbf{DermNetNZ} \cite{dermnet2025} dataset were selected, with 700 high-quality images (100 for each disease). Each image is paired with a Question–Answer (Q\&A) label, as shown in Figure~\ref{fig:dataset_example}.

\vspace{0.1cm}

\subsubsection{\textbf{Image and Conversation Dataset for Fine-tuning}}

Our SFT dataset trains the model to recognize diseases and explain diagnostic reasoning in a clear, human-understandable manner. Each instance includes a dermatological image, a user question, and a detailed, multi-part ground-truth answer. The answer reflects a step-by-step thought process of a dermatologist, examining features like color, texture, and shape to arrive at a diagnosis. The Q\&A pairs in Figure~\ref{fig:dataset_example} demonstrate typical data points in this dataset.

\vspace{0.1cm}

\subsubsection{\textbf{Preference Conversation Dataset for DPO}}

To ensure factual, knowledge-grounded reasoning, we constructed a preference dataset \textit{online} during DPO training (Figure~\ref{fig:dataset_example}). At each step, the fine-tuned model generates two responses: one directly from the base model and another using KG-RAG. These are compared on quality, accuracy, and factual correctness. In our setup, the KG-RAG output is labeled as \textbf{chosen} and the base output as \textbf{rejected}, forming a consistent preference dataset for DPO optimization. This process enables the model to internalize clinically reliable patterns, enhancing factual accuracy.

\subsection{\textbf{Training Stages}}
\subsubsection{\textbf{Stage-1: Reinforcement Learning with GRPO++}}

In clinical practice, dermatologists diagnose through structured reasoning: identifying findings, forming hypotheses, and weighing them with context. Stage 1 aims to instill this stepwise logic in a vision-language model, avoiding shallow predictions. Grouped Relative Policy Optimization (GRPO) serves as a basis: for each prompt, multiple responses are generated, compared within a group, and reinforced if stronger than peers—mirroring differential diagnosis. Yet, \textbf{GRPO has two weaknesses}. If all responses are wrong but receive different rewards, the least wrong is still reinforced (error reinforcement). If all are wrong with identical rewards, variance collapses and learning stalls (advantage collapse).

\medskip

\begin{figure*}[t!]
    \centering
    \begin{subfigure}{0.48\textwidth}
        \includegraphics[width=\linewidth]{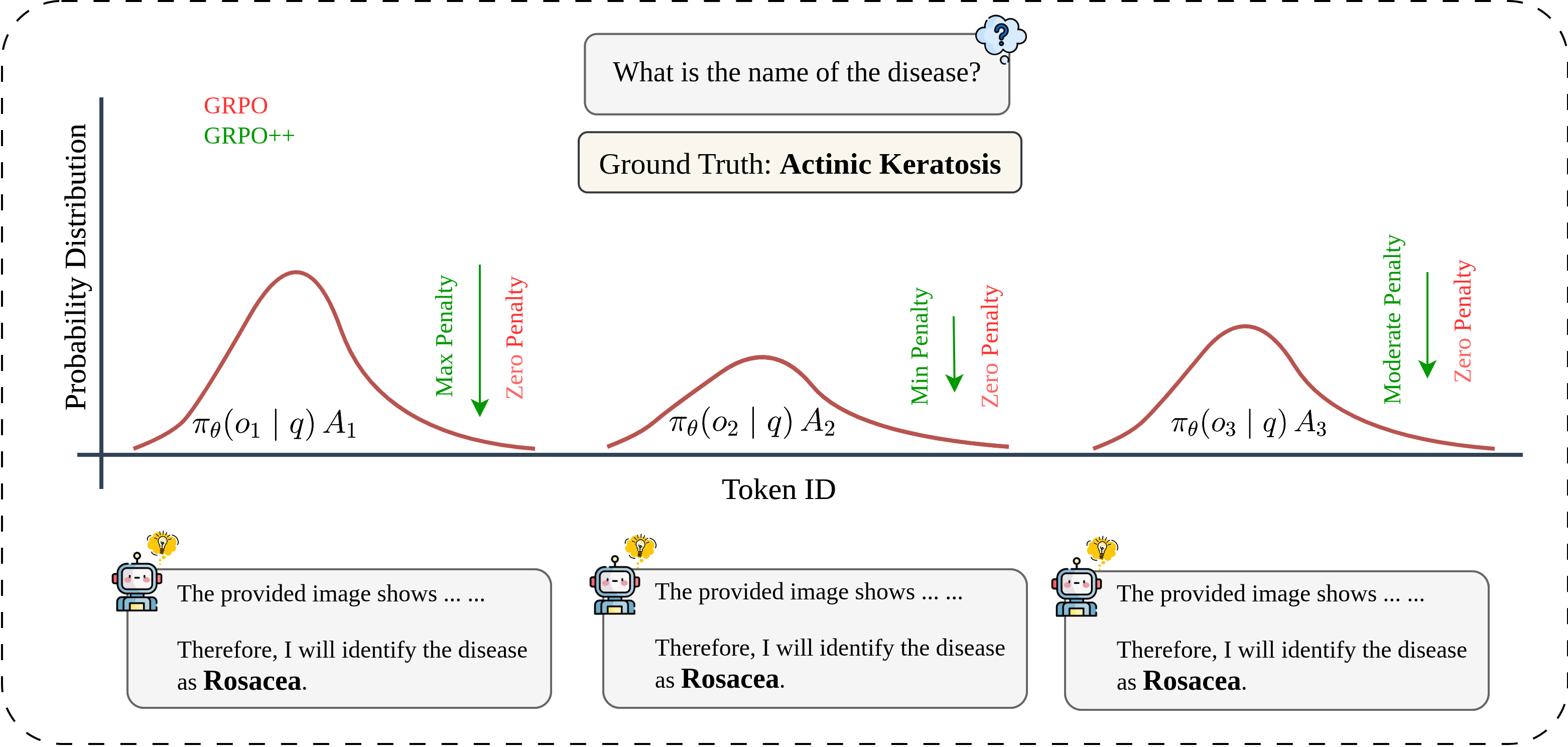}
        \caption{Failure mode \ref{failure mode 1} for GRPO in a low-generation setting: repeated identical outputs within a group cause the intra-group advantage to collapse. This is fixed by GRPO++ in \ref{corrected}}
        \label{GRPO++ vs GRPO 1}
    \end{subfigure}
    \hfill
    \begin{subfigure}{0.48\textwidth}
        \includegraphics[width=\linewidth]{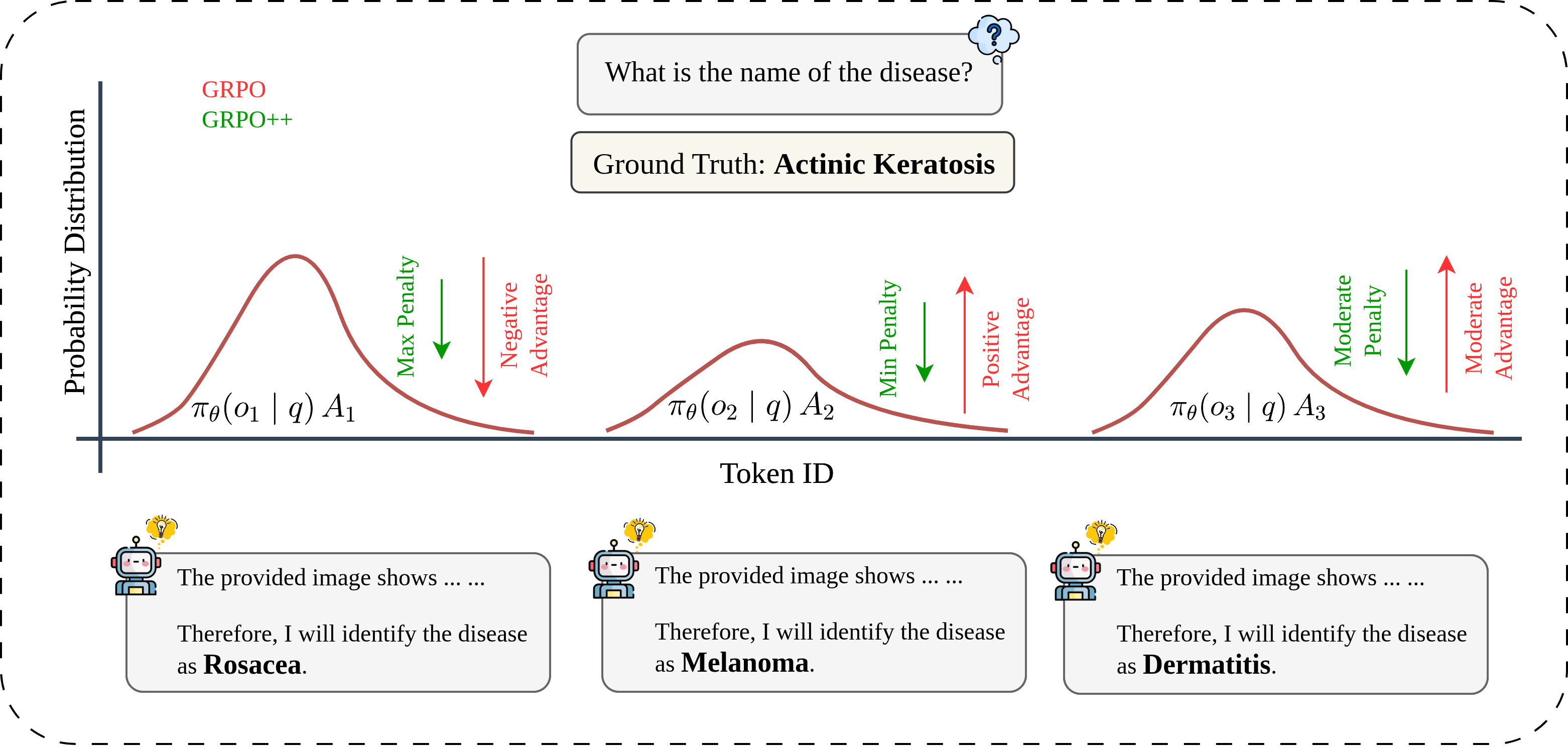}
        \caption{Failure mode \ref{failure mode 2} for GRPO in a low-generation setting: repeated identical outputs within a group cause the intra-group advantage to collapse. This is fixed by GRPO++ in \ref{corrected}}
        \label{GRPO++ vs GRPO 2}
    \end{subfigure}
    \caption{Failure mode comparison across GRPO and GRPO++.}
    \label{GRPO vs GRPO++}
\end{figure*}

To address this, we propose \textbf{GRPO++}, a variant of GRPO equipped with a \textit{confidence-aware advantage function}. When at least one response is correct, GRPO++ reduces to standard GRPO, promoting stronger candidates through normalized advantages. However, when all responses are incorrect, relative comparisons provide no learning signal. In such cases, GRPO++ switches to an absolute, confidence-weighted penalty that penalizes high-probability incorrect responses more heavily than uncertain ones. This mechanism prevents both error reinforcement and learning collapse, enabling progress even in low-diversity or all-wrong settings. As shown in Figure.~\ref{GRPO++ vs GRPO 1}, when all responses are identical and wrong, GRPO assigns zero advantage to every response, stalling the training of small language models that often repeat the same outputs. In contrast, GRPO++ imposes stronger penalties on overconfident but incorrect responses while assigning lighter penalties to less confident ones. Figure.~\ref{GRPO++ vs GRPO 2} further illustrates that GRPO tends to reinforce suboptimal responses that are less penalized, causing smaller models to get stuck repeating them. By comparison, GRPO++ discourages such behaviors and more effectively drives the model toward generating correct responses.

\medskip

\noindent \textbf{Confidence-Aware Advantage Function:}  

Let $\mathcal{C} = \{i : r_i \geq \tau\}$ denote responses above threshold $\tau$ (set to $0$ in our case). We then define:
\begin{equation}
\hat{A}^{\text{CA}}_{i,t} =
\begin{cases}
\frac{R_{i,t} - \bar{R}}{\sigma_R + \varepsilon}, & \text{if } |\mathcal{C}| \geq 1 \\[0.3em]
-\beta \cdot \frac{\ell_i - \ell_{\min}}{\ell_{\max} - \ell_{\min} + \varepsilon} - \gamma, & \text{if } |\mathcal{C}| = 0
\end{cases}
\label{eq:ca_advantage_final}
\end{equation}
where $R_{i,t}$ is reward-to-go, $\bar{R}, \sigma_R$ are mean and standard deviation of rewards, and $\ell_i = \sum_t \log \pi_{\theta_{\text{old}}}(o_{i,t}\mid q,o_{i,<t})$. The terms $\ell_{\min}, \ell_{\max}$ denote group extremes. Coefficients $\beta>0,\gamma>0$ control penalty strength, and $\varepsilon$ prevents division by zero.

\medskip

\noindent \textbf{GRPO++ objective:}  
\begin{equation}
\begin{aligned}
\mathcal{J}_{\text{GRPO++}} &= \mathbb{E}_{q,\{o_i\}}\Bigg[\frac{1}{m}\sum_{i=1}^m\frac{1}{|o_i|}\sum_{t=1}^{|o_i|} \\
&\quad \min\{\rho_{i,t}\hat{A}^{\text{CA}}_{i,t}, \text{clip}(\rho_{i,t})\hat{A}^{\text{CA}}_{i,t}\}\Bigg],
\end{aligned}
\label{eq:cagrpo}
\end{equation}
where $m$ is the number of responses, $|o_i|$ their length, $\rho_{i,t}$ the importance ratio, and $\text{clip}(\rho) = \text{clip}(\rho,1-\epsilon,1+\epsilon)$ defines the trust region.

This ensures GRPO++ yields informative updates even when all responses fail. Section~\ref{theoretical analysis} derives this formally. Algorithm~\ref{alg:cagrpo} summarizes the training loop: responses are generated, rewards computed, and advantages assigned. If at least one response is correct, GRPO updates apply; otherwise, the confidence-aware penalty is triggered. Applied to Qwen2-VL-2B and Qwen2.5-VL-3B, this produces the Reasoning-Aligned VLM (Figure~\ref{fig:training_flow}) specialized in visual disease detection.

\begin{algorithm}[t]
\caption{Group Relative Policy Optimization ++ (GRPO++)}
\label{alg:cagrpo}
\begin{algorithmic}[1]
\State \textbf{Input:} policy $\pi_\theta$, reward model $r_\phi$, dataset $\mathcal{D}$
\State \textbf{Parameters:} $\epsilon$, $\beta$, $\gamma$, threshold $\tau$
\For{iteration $l = 1, \ldots, I$}
    \State $\pi_{\text{ref}} \gets \pi_\theta$
    \For{step $s = 1, \ldots, M$}
        \State Sample batch $\mathcal{D}_b \sim \mathcal{D}$
        \State $\pi_{\theta_{\text{old}}} \gets \pi_\theta$
        \For{prompt $q \in \mathcal{D}_b$}
            \State $\{o_i\}_{i=1}^m \sim \pi_{\theta_{\text{old}}}(\cdot|q)$
            \State $\{r_i\}_{i=1}^m \gets r_\phi(\{o_i\})$
            \State $\mathcal{C} \gets \{i : r_i \geq \tau\}$
            \If{$|\mathcal{C}| \geq 1$}
                \State Compute standard GRPO advantages
            \Else
                \State Apply confidence-aware penalty (Eq.~\ref{eq:ca_advantage_final})
            \EndIf
        \EndFor
        \For{PPO step $t = 1, \ldots, T$}
            \State $\theta \gets \theta + \alpha \nabla_\theta \mathcal{J}_{\text{GRPO++}}$
        \EndFor
    \EndFor
\EndFor
\State \textbf{Return:} optimized policy $\pi_\theta$
\end{algorithmic}
\end{algorithm}

\vspace{0.1cm}
\subsubsection{\textbf{Stage-2: Supervised Fine-Tuning with Image Conversation Dataset}}
\label{Stage 2: Dermatological Concept & Knowledge Alignment}

Stage~1 yields the \textbf{Reasoning Aligned VLM}, tuned for visual reasoning (Fig.~\ref{fig:training_flow}). While effective at image-based detection, it lacks broader clinical knowledge needed to discuss causes or treatments. To bridge this, we apply \textbf{Supervised Fine-Tuning (SFT)} on an \textbf{Image Conversation Dataset}, producing the \textbf{Conversation Tuned VLM}. This equips the model with clinically grounded conversational ability, enabling it to explain diagnoses, discuss etiologies, and suggest treatments.

\medskip
\subsubsection{\textbf{Stage-3: Improving Diagnostic Accuracy via Knowledge Graphs and Preference Tuning}}

Dermatologists consult texts, research databases, and guidelines for complex cases, including rare conditions. To reduce hallucination and factual errors, we adopt a Knowledge Graph-based Retrieval-Augmented Generation (KG-RAG). Relevant triples (symptoms, causes, treatments) are retrieved and integrated, grounding responses in validated facts.

However, continual retrieval increases overhead due to long contexts. To emulate how dermatologists internalize knowledge, we refine Stage-2 with \textbf{Direct Preference Optimization (DPO)}. An online preference dataset is built from paired outputs: one with KG-RAG (\textbf{chosen}) and one without (\textbf{rejected}). Training with DPO aligns the model to prefer knowledge-grounded answers, internalizing medical patterns while reducing reliance on retrieval. The final model, \textbf{DermIQ-VLM}, delivers accurate and efficient responses, reflecting a dermatologist’s balance of expertise and practicality (Figure~\ref{fig:training_flow}).

\section{Theoretical Analysis}
\label{theoretical analysis}
In this section, we provide rigorous theoretical foundations for GRPO++, demonstrating how our confidence-aware modifications address fundamental limitations of standard GRPO while preserving convergence guarantees. We identify two critical failure modes in standard GRPO and establish that our proposed method systematically overcomes these limitations.

\subsection{\textbf{Failure Modes of Standard GRPO}}

When training models for complex reasoning tasks in low-resource settings—such as generating only a small number of responses per prompt or using smaller language models that tend to produce similar outputs—standard GRPO suffers from two critical failure modes that hinder learning: (i) gradient vanishing due to low response diversity, and (ii) systematic reinforcement of suboptimal behaviors.

\subsubsection{Gradient Vanishing Under Low Diversity}
\label{failure mode 1}

Consider response set $\mathcal{O} = \{o_1, ..., o_m\}$ with rewards $\mathcal{R} = \{r_1, ..., r_m\}$. In low diversity regimes—common with small models (<4B parameters) or limited sampling—responses converge such that $|r_i - r_j| < \delta$ for small $\delta > 0$.

The standard GRPO advantage function:
\begin{equation}
\hat{A}^{\text{GRPO}}_{i,t} = \frac{r_i - \bar{r}}{\sigma_r + \varepsilon}
\end{equation}
where $\bar{r} = \frac{1}{m}\sum_{j} r_j$ and $\sigma_r = \sqrt{\frac{1}{m}\sum_{j}(r_j - \bar{r})^2}$.

When $r_i \approx r_c$ for all $i$:
\begin{align}
\bar{r} &\approx r_c \\
\sigma_r &\approx 0 \\
\hat{A}^{\text{GRPO}}_{i,t} &\approx \frac{r_c - r_c}{0 + \varepsilon} = 0
\end{align}

This yields vanishing gradients:
\begin{equation}
\nabla_\theta \mathcal{J}_{\text{GRPO}} = \sum_{i,t} \nabla_\theta \log \pi_\theta(o_{i,t}|x) \cdot 0 = 0
\end{equation}

Critically, this occurs regardless of whether $r_c$ represents high or low quality, preventing improvement when converged to suboptimal solutions.

\subsubsection{Error Reinforcement Problem}
\label{failure mode 2}
Even with diversity, when all responses are suboptimal ($r_i < \tau$ for all $i$), the zero-sum property ensures:
\begin{equation}
\sum_{i=1}^m (r_i - \bar{r}) = 0 \implies \exists i^* : r_{i^*} > \bar{r}
\end{equation}

Response $i^*$ receives positive reinforcement despite being suboptimal:
\begin{equation}
\hat{A}^{\text{GRPO}}_{i^*,t} = \frac{r_{i^*} - \bar{r}}{\sigma_r + \varepsilon} > 0, \quad r_{i^*} < \tau
\end{equation}

This systematically reinforces inadequate responses, potentially causing convergence on suboptimal solutions.

\subsection{\textbf{GRPO++ Solution}}

GRPO++ introduces confidence scores based on log-likelihood:
\begin{equation}
\ell_i = \sum_{t=1}^{|o_i|} \log \pi_{\theta_{\text{old}}}(o_{i,t}|q,o_{i,<t})
\end{equation}

For confidence set $\mathcal{C} = \{i : r_i \geq \tau\}$, when $|\mathcal{C}|=0$ (all suboptimal):
\begin{equation}
\hat{A}^{\text{CA}}_{i,t} = -\beta \cdot \frac{\ell_i - \ell_{\min}}{\ell_{\max} - \ell_{\min} + \varepsilon} - \gamma
\end{equation}
where $\beta, \gamma > 0$ are penalty parameters. This ensures $\hat{A}^{\text{CA}}_{i,t} < 0$ for all $i,t$.


\subsection{Gradient Analysis of GRPO++}
\label{corrected}

We analyze GRPO++ gradient properties under the suboptimal regime where all responses fail the quality threshold ($|\mathcal{C}| = 0$).

\textbf{Theorem 1 (Gradient Vanishing Condition).} Under conditions: (i) all responses suboptimal, (ii) $\ell_{\max} > \ell_{\min}$, GRPO++ gradients vanish iff $\sum_{i=1}^m (\gamma + \beta w_i) s_i = 0$.

\textit{Proof.} The confidence-aware advantage is:
\begin{equation}
\hat{A}_i = -\gamma - \beta w_i
\end{equation}
where $w_i = (\ell_i - \ell_{\min})/(\ell_{\max} - \ell_{\min} + \varepsilon)$.

The gradient becomes:
\begin{equation}
\nabla_\theta \mathcal{J}_{\text{GRPO++}} = -\frac{1}{m}\sum_{i=1}^m (\gamma + \beta w_i) s_i
\end{equation}

Vanishing requires the weighted sum of score functions to equal zero. Unlike standard GRPO, this condition depends on both confidence weights $w_i$ and policy gradients $s_i$.

\textbf{Theorem 2 (Gradient Bounds).} Let $G_{\max} = \max_i \|s_i\|$, $G_{\min} = \min_i \|s_i\|$. Under no perfect cancellation:
\begin{equation}
\left(\gamma + \frac{\beta}{m}\right) G_{\min} \leq \|\nabla_\theta \mathcal{J}_{\text{GRPO++}}\| \leq \left(\gamma + \frac{(m-1)\beta}{m}\right) G_{\max}
\end{equation}

\textit{Proof.} Apply triangle inequality: $\|\nabla_\theta \mathcal{J}_{\text{GRPO++}}\| \leq \frac{1}{m}\sum_{i=1}^m |\hat{A}_i| \|s_i\|$.

Since $|\hat{A}_i| = \gamma + \beta w_i$, the constraint $\ell_{\max} > \ell_{\min}$ prevents all $w_i = 1$ simultaneously.

\textit{Upper bound:} Extremal case with $(m-1)$ responses having $w_i = 1$, one with $w_i = 0$:
\begin{align}
\|\nabla\| &\leq \frac{(m-1)(\gamma + \beta) + \gamma}{m} G_{\max} \\
&= \left(\gamma + \frac{(m-1)\beta}{m}\right) G_{\max}
\end{align}

\textit{Lower bound:} Extremal case with one response having $w_i = 1$, $(m-1)$ with $w_i = 0$:
\begin{align}
\|\nabla\| &\geq \frac{(\gamma + \beta) + (m-1)\gamma}{m} G_{\min} \\
&= \left(\gamma + \frac{\beta}{m}\right) G_{\min}
\end{align}

The derived bounds are \textbf{provably tight}, as they can be achieved under specific extremal conditions. The upper bound is attained when the confidence distribution places $(m-1)$ responses at $\ell_{\max}$ and one at $\ell_{\min}$, with all score vectors aligned constructively. Conversely, the lower bound is realized when one response is at $\ell_{\max}$ and the remaining $(m-1)$ at $\ell_{\min}$, with minimal destructive interference. Moreover, the $\tfrac{\beta}{m}$ corrections emerge from the structural constraint that it is fundamentally impossible for all $w_i$ to equal $1$ simultaneously.

\begin{figure*}[t!]
    \centering
    \begin{subfigure}{0.48\textwidth}
        \includegraphics[width=\linewidth]{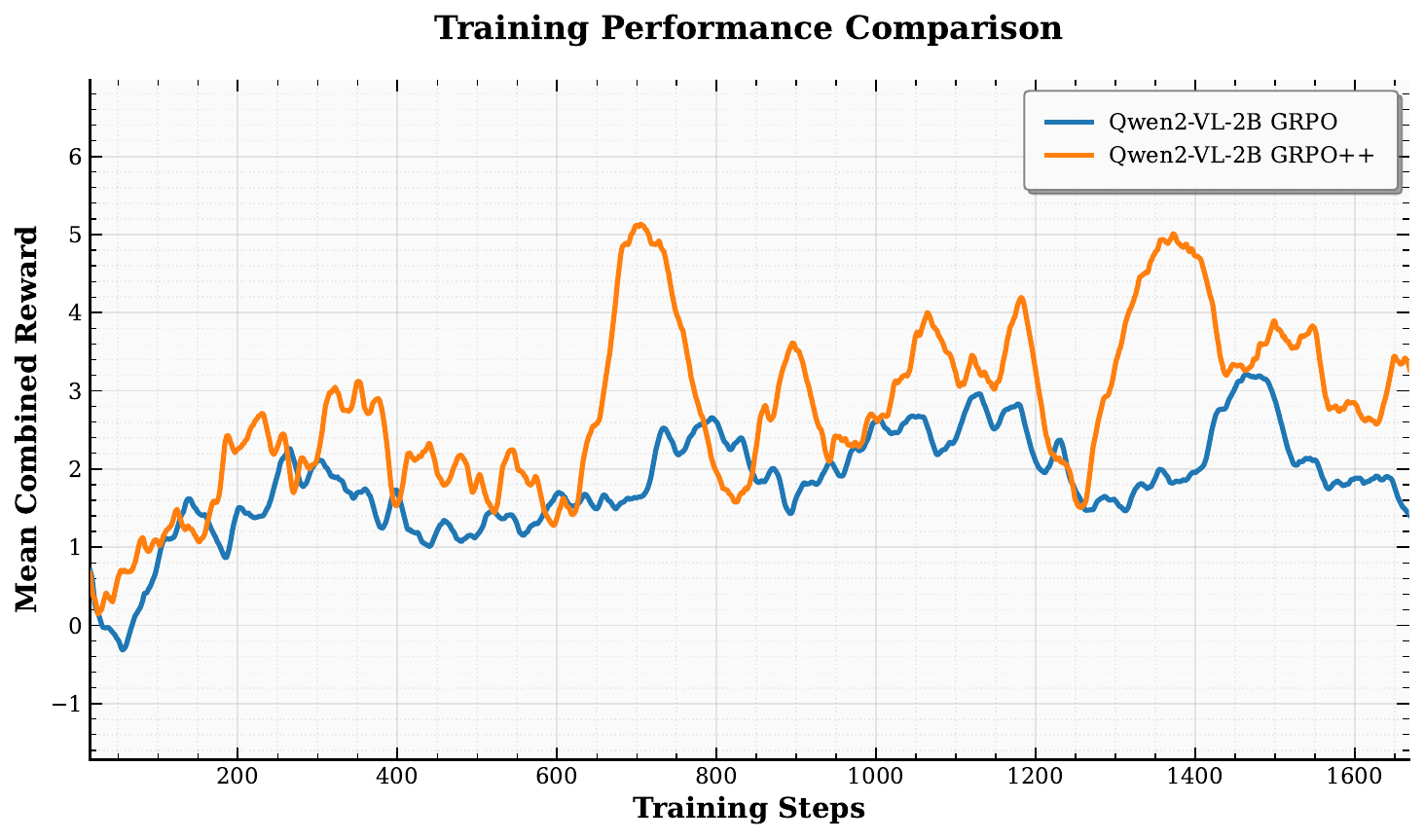}
        \caption{GRPO vs GRPO++ (Qwen2-VL-2B)}
        \label{fig:comparison_v1}
    \end{subfigure}
    \hfill
    \begin{subfigure}{0.48\textwidth}
        \includegraphics[width=\linewidth]{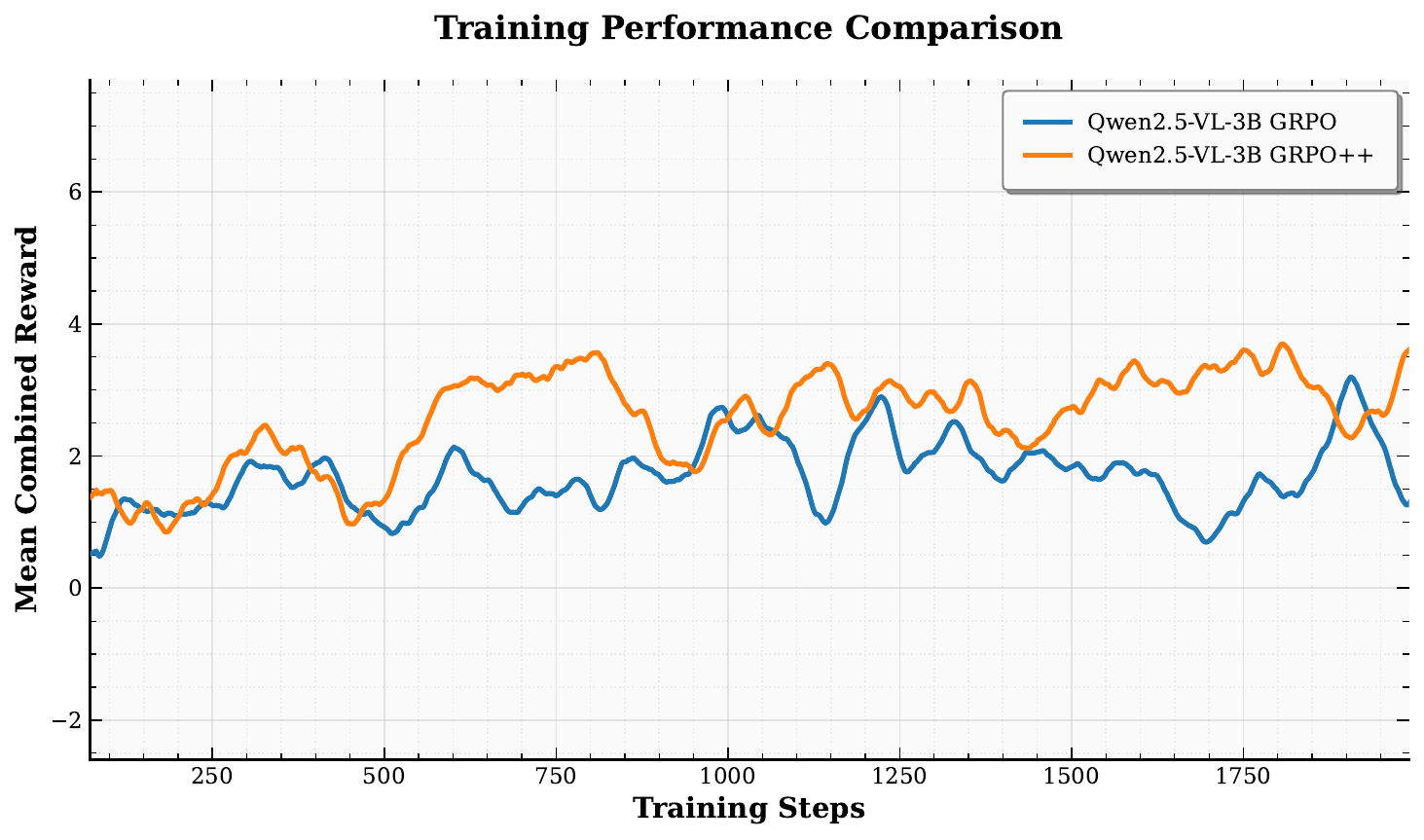}
        \caption{GRPO vs GRPO++ (Qwen2.5-VL-3B)}
        \label{fig:comparison_v2}
    \end{subfigure}
    \caption{Reward comparison across GRPO and GRPO++.}
    \label{fig:comparison}
\end{figure*}

\section{Experiments and Results}\label{sec5}
We built a custom benchmark of 138 unseen image pairs~\cite{ismam_nur_swapnil_2025} from DermNetNZ~\cite{dermnet2025}, with $\sim$20 images per class, to evaluate dermatological disease detection.

\subsection{\textbf{Experimental Setup}}

For \textbf{Stage-1: GRPO++}, training used two 15GB T4 GPUs with 4-bit quantization and LoRA to reduce memory and computation. LoRA (rank=32, $\alpha$=64, dropout=0.05) targeted \texttt{q\_proj}, \texttt{k\_proj}, \texttt{v\_proj}, and \texttt{o\_proj}. Training ran for $\sim$1700 steps (10 epochs) with learning rate $1e{-5}$, batch size 1 (4 grad accumulation), and temperature 0.9. Generation parameters were set to 3 for both 2B and 3B models. For \textbf{Stage-2: Fine-Tune} and \textbf{Stage-3: DPO} (Figure~\ref{fig:training_flow}), 4-bit quantization and LoRA were again used (rank=8, $\alpha$=16, dropout=0.05), expanding targets to include \texttt{down\_proj}, \texttt{up\_proj}, \texttt{gate\_proj}, etc. Training used $1e{-5}$ learning rate, 2 epochs, batch size 1 (2 grad accumulation), AdamW (weight decay=0.01), linear scheduler (warmup=0.03), and gradient clipping (0.3).

\subsection{\textbf{Baselines and Metrics}}

DermIQ-VLM was compared against Qwen2.5-VL-2B/3B-Instruct. For detection, answers between \texttt{<answer>}...\texttt{</answer>} were matched to ground truth; majority voting aggregated predictions for robustness.

Conversation quality was judged by Grok and GPT-4, checking adherence to \texttt{<thinking>}...\texttt{</thinking>} and \texttt{<answer>}...\texttt{</answer>} formats. Metrics included \textbf{factual accuracy} (disease correctness), \textbf{relevance} (query alignment), and \textbf{completeness} (coverage of reasoning and diagnosis).

\subsection{\textbf{Detected Disease Reward:}}

This reward guides the model toward accurate, clinically safe diagnoses by combining general reward constants (Table~\ref{tab:tab1}) with a severity-based penalty matrix (Table~\ref{tab:comprehensive_reward_structure}). Correct predictions earn a base reward, while errors incur penalties scaled by clinical risk. For example, misclassifying Melanoma or cancerous disease as Dermatitis or inflamatory disease yields a severe penalty of \(-5.0\), whereas confusing Actinic Keratosis with Seborrheic Keratosis incurs only \(-2.0\). Additional rules handle invalid predictions, unknown ground truth, and default mismatches.

\begin{table}[htbp]
\centering
\resizebox{\linewidth}{!}{%
\begin{tabular}{l c}
\toprule
\textbf{Parameter Description} & \textbf{Value} \\
\midrule
Base reward for correct disease identification  & \(+10.0\) \\
Penalty for failing to output a valid disease  & \(-5.0\) \\
Penalty if ground truth label is unknown/invalid & \(-0.5\) \\
Default penalty for unlisted misclassifications  & \(-2.5\) \\
\bottomrule
\end{tabular}%
}
\caption{General reward and penalty constants.}
\label{tab:tab1}
\end{table}

\begin{table}[htbp]
\centering
\resizebox{\linewidth}{!}{%
\begin{tabular}{l c c c c c c c}
\toprule
\textbf{True Disease} & AK & BCC & Derm. & Mel. & Psor. & Ros. & SK \\
\midrule
AK   & N/A  & -1.0  & -3.0  & -1.5  & -3.0  & -3.0  & -2.0  \\
BCC  & -1.5 & N/A   & -4.0  & -2.0  & -4.0  & -4.0  & -3.0  \\
Derm.& -2.5 & -3.0  & N/A   & -3.5  & -0.5  & -0.7  & -2.5  \\
Mel. & -3.0 & -2.5  & -5.0  & N/A   & -5.0  & -5.0  & -4.0  \\
Psor.& -2.5 & -3.0  & -0.5  & -3.5  & N/A   & -0.8  & -2.5  \\
Ros. & -2.5 & -3.0  & -0.7  & -3.5  & -0.8  & N/A   & -2.5  \\
SK   & -1.0 & -2.0  & -1.5  & -3.0  & -1.5  & -1.5  & N/A   \\
\bottomrule
\end{tabular}%
}
\caption{Severity-based penalty matrix for disease misclassifications. Abbreviations: AK = Actinic Keratosis, BCC = Basal Cell Carcinoma, Derm. = Dermatitis, Mel. = Melanoma, Psor. = Psoriasis, Ros. = Rosacea, SK = Seborrheic Keratosis.}

\label{tab:comprehensive_reward_structure}
\end{table}

\subsection{\textbf{Performance of Disease Detection with Reasoning}}

We compared pretrained VLMs with GRPO-tuned and DermIQ-VLM variants. Single-shot results (Table~\ref{tab:single_shot_summary}) show pretrained models perform poorly, especially the 2B backbone. GRPO tuning improves accuracy, while DermIQ-VLM achieves the strongest gains across both model sizes.

\begin{table}[h]
\centering
\resizebox{\linewidth}{!}{%
\begin{tabular}{l l c c c}
\toprule
\textbf{Type} & \textbf{Model} & \textbf{F1 (\%)} & \textbf{Prec. (\%)} & \textbf{Rec. (\%)} \\
\midrule
Pretrained & Qwen2-VL-2B & 7.35 & 13.47 & 15.94 \\
           & Qwen2.5-VL-3B & 19.93 & 27.62 & 21.01 \\
\midrule
GRPO-tuned & Qwen2-VL-2B & 33.97 & 42.98 & 39.05 \\
           & Qwen2.5-VL-3B & 42.31 & 42.36 & 45.69 \\
\midrule
DermIQ-VLM & Qwen2-VL-2B & \textbf{41.28} & \textbf{47.42} & \textbf{40.32} \\
           & Qwen2.5-VL-3B & \textbf{45.74} & \textbf{48.73} & \textbf{47.58} \\
\bottomrule
\end{tabular}%
}
\caption{Single-shot evaluation performance.}
\label{tab:single_shot_summary}
\end{table}

\begin{table}[h]
\centering
\resizebox{\linewidth}{!}{%
\begin{tabular}{l l c c c}
\toprule
\textbf{Type} & \textbf{Model} & \textbf{F1 (\%)} & \textbf{Prec. (\%)} & \textbf{Rec. (\%)} \\
\midrule
Pretrained & Qwen2-VL-2B & 19.02 & 21.95 & 19.29 \\
           & Qwen2.5-VL-3B & 24.85 & 39.45 & 27.73 \\
\midrule
GRPO-tuned & Qwen2-VL-2B & 42.31 & 42.36 & 45.69 \\
           & Qwen2.5-VL-3B & 48.20 & 51.11 & 51.05 \\
\midrule
DermIQ-VLM & Qwen2-VL-2B & \textbf{47.57} & \textbf{58.99} & \textbf{48.12} \\
           & Qwen2.5-VL-3B & \textbf{51.38} & \textbf{57.32} & \textbf{52.90} \\
\bottomrule
\end{tabular}%
}
\caption{Majority voting evaluation performance.}
\label{tab:majority_voting_summary}
\end{table}

Majority voting (Table~\ref{tab:majority_voting_summary}) boosts all models, with DermIQ-VLM consistently leading. Improvements are most notable in challenging diseases like Seborrheic Keratosis, where baselines fail. Detailed per-disease reports (Tables~\ref{tab:single_shot_detailed_per_disease},~\ref{tab:majority_voting_detailed_per_disease}) confirm DermIQ-VLM excels on cancers (AK, BCC, Melanoma) and rare classes (SK), while GRPO remains competitive on Dermatitis, Psoriasis, and Rosacea but fails miserably on Seborrheic Keratosis (SK). Together, the results show GRPO++ narrows performance gaps across backbones and enhances reliability in both single-shot and aggregated settings. Across both model backbones, \textbf{GRPO++} (orange) consistently outperforms \textbf{GRPO} (blue), delivering higher combined rewards and a steady upward training trend which can be seen in Figure~\ref{fig:comparison}. This advantage persists across model sizes, with Qwen2.5-VL-3B achieving stronger overall rewards than Qwen2-VL-2B. Beyond improving per-disease performance in the single-shot setting, GRPO++ also amplifies ensemble-style gains. Its robustness is most evident in difficult cases like Seborrheic Keratosis, where baseline models collapse, while GRPO-tuned models sustain performance on conditions such as Dermatitis, Psoriasis, and Rosacea.

\begin{table}[h]
\centering
\resizebox{\linewidth}{!}{%
\begin{tabular}{l ccc ccc ccc ccc}
\toprule
\multirow{3}{*}{\textbf{Disease}} 
 & \multicolumn{6}{c}{\textbf{Pretrained VLMs}} 
 & \multicolumn{3}{c}{\textbf{GRPO-tuned VLM}} 
 & \multicolumn{3}{c}{\textbf{DermIQ-VLM}} \\
\cmidrule(lr){2-7} \cmidrule(lr){8-10} \cmidrule(lr){11-13}
 & \multicolumn{3}{c}{\textbf{Qwen2-VL-2B}} 
 & \multicolumn{3}{c}{\textbf{Qwen2.5-VL-3B}} 
 & \multicolumn{3}{c}{\textbf{Qwen2.5-VL-3B}} 
 & \multicolumn{3}{c}{\textbf{Qwen2.5-VL-3B}} \\
\cmidrule(lr){2-4} \cmidrule(lr){5-7} \cmidrule(lr){8-10} \cmidrule(lr){11-13}
 & P & R & F1 & P & R & F1 & P & R & F1 & P & R & F1 \\
\midrule
AK  & 0.20 & 0.05 & 0.08 & 0.13 & 0.41 & 0.20 & 0.22 & 0.31 & 0.26 & 0.55 & 0.30 & 0.39 \\
BCC & 0.13 & 0.05 & 0.07 & 0.21 & 0.41 & 0.28 & 0.42 & 0.63 & 0.50 & 0.47 & 0.79 & 0.59 \\
DER & 0.15 & 0.95 & 0.27 & 0.00 & 0.00 & 0.00 & 0.26 & 0.33 & 0.29 & 0.22 & 0.12 & 0.15 \\
MEL & 0.50 & 0.06 & 0.10 & 0.71 & 0.28 & 0.40 & 0.69 & 0.61 & 0.65 & 0.60 & 0.75 & 0.67 \\
PSO & 0.00 & 0.00 & 0.00 & 0.10 & 0.06 & 0.07 & 0.67 & 0.29 & 0.40 & 0.50 & 0.21 & 0.30 \\
ROS & 0.00 & 0.00 & 0.00 & 0.83 & 0.33 & 0.48 & 0.67 & 0.90 & 0.77 & 0.82 & 0.74 & 0.78 \\
SK  & 0.00 & 0.00 & 0.00 & 0.00 & 0.00 & 0.00 & 0.00 & 0.00 & 0.00 & 0.24 & 0.37 & 0.29 \\
\bottomrule
\end{tabular}%
}
\caption{Single Shot Evaluation: Precision (P), Recall (R), and F1-Score (F1) per disease for each model (mostly Qwen2.5-VL-3B).}
\label{tab:single_shot_detailed_per_disease}
\end{table}

\begin{table}[h]
\centering
\resizebox{\linewidth}{!}{%
\begin{tabular}{l ccc ccc ccc ccc}
\toprule
\multirow{3}{*}{\textbf{Disease}} 
 & \multicolumn{6}{c}{\textbf{Pretrained VLMs}} 
 & \multicolumn{3}{c}{\textbf{GRPO-tuned VLM}} 
 & \multicolumn{3}{c}{\textbf{DermIQ-VLM}} \\
\cmidrule(lr){2-7} \cmidrule(lr){8-10} \cmidrule(lr){11-13}
 & \multicolumn{3}{c}{\textbf{Qwen2-VL-2B}} 
 & \multicolumn{3}{c}{\textbf{Qwen2.5-VL-3B}} 
 & \multicolumn{3}{c}{\textbf{Qwen2.5-VL-3B}} 
 & \multicolumn{3}{c}{\textbf{Qwen2.5-VL-3B}} \\
\cmidrule(lr){2-4} \cmidrule(lr){5-7} \cmidrule(lr){8-10} \cmidrule(lr){11-13}
 & P & R & F1 & P & R & F1 & P & R & F1 & P & R & F1 \\
\midrule
AK  & 0.40 & 0.10 & 0.16 & 0.19 & 0.76 & 0.31 & 0.22 & 0.31 & 0.26 & 0.50 & 0.60 & 0.55 \\
BCC & 0.38 & 0.15 & 0.21 & 0.25 & 0.41 & 0.31 & 0.42 & 0.63 & 0.50 & 0.45 & 0.50 & 0.48 \\
DER & 0.18 & 0.80 & 0.29 & 0.25 & 0.06 & 0.09 & 0.26 & 0.33 & 0.29 & 0.45 & 0.62 & 0.52 \\
MEL & 0.33 & 0.10 & 0.15 & 0.75 & 0.17 & 0.27 & 0.69 & 0.61 & 0.65 & 0.67 & 0.67 & 0.67 \\
PSO & 0.50 & 0.15 & 0.23 & 0.56 & 0.29 & 0.38 & 0.67 & 0.29 & 0.40 & 0.77 & 0.50 & 0.61 \\
ROS & 0.28 & 0.25 & 0.26 & 0.80 & 0.27 & 0.40 & 0.67 & 0.90 & 0.77 & 0.65 & 0.65 & 0.65 \\
SK  & 0.17 & 0.05 & 0.08 & 0.00 & 0.00 & 0.00 & 0.00 & 0.00 & 0.00 & 0.69 & 0.45 & 0.55 \\
\bottomrule
\end{tabular}%
}
\caption{Majority Voting Evaluation: Precision (P), Recall (R), and F1-Score (F1) per disease for each model (mostly Qwen2.5-VL-3B).}
\label{tab:majority_voting_detailed_per_disease}
\end{table}

\begin{table}[h]
\small
\centering
\resizebox{\linewidth}{!}{%
\begin{tabular}{ll*{6}{S}}
\toprule
\multirow{2}{*}{\textbf{Type}} & \multirow{2}{*}{\textbf{Topic}} 
& \multicolumn{2}{c}{\textbf{Accuracy}} 
& \multicolumn{2}{c}{\textbf{Relevance}} 
& \multicolumn{2}{c}{\textbf{Completeness}} \\
\cmidrule(lr){3-4} \cmidrule(lr){5-6} \cmidrule(lr){7-8}
& & {\textbf{Grok}} & {\textbf{GPT-4}} 
& {\textbf{Grok}} & {\textbf{GPT-4}} 
& {\textbf{Grok}} & {\textbf{GPT-4}} \\
\midrule
\rowcolor{gray!15} \multicolumn{8}{c}{\textbf{Backbone Model: Qwen2-VL-2B}} \\
\midrule
\multirow{4}{*}{Pretrained} 
& Treatment     & 4.9 & 4.3 & 5.1 & 4.4 & 4.7 & 4.0 \\
& Causes       & 3.5 & 3.0 & 4.7 & 4.1 & 4.2 & 3.7 \\
& Demographics & 4.6 & 4.1 & 4.4 & 3.9 & 4.5 & 4.0 \\
& Features     & 6.1 & 5.3 & 5.9 & 5.1 & 5.6 & 4.8 \\
& \textbf{Average} & \textbf{4.78} & \textbf{4.18} & \textbf{5.03} & \textbf{4.38} & \textbf{4.75} & \textbf{4.13} \\
\midrule
\multirow{4}{*}{\shortstack{Conversation \\ Tuned}} 
& Treatment     & 6.3 & 5.7 & 6.1 & 5.5 & 6.0 & 5.4 \\
& Causes       & 6.6 & 6.0 & 6.2 & 5.7 & 5.8 & 5.2 \\
& Demographics & 5.5 & 5.0 & 4.7 & 4.2 & 5.1 & 4.6 \\
& Features     & 6.8 & 6.2 & 6.5 & 5.9 & 6.0 & 5.5 \\
& \textbf{Average} & \textbf{6.30} & \textbf{5.73} & \textbf{5.88} & \textbf{5.33} & \textbf{5.73} & \textbf{5.18} \\
\midrule
\multirow{4}{*}{DermIQ-VLM} 
& Treatment     & 7.2 & 6.6 & 6.9 & 6.3 & 7.0 & 6.4 \\
& Causes       & 7.0 & 6.4 & 6.8 & 6.2 & 6.6 & 6.0 \\
& Demographics & 6.7 & 6.1 & 6.4 & 5.9 & 6.5 & 6.0 \\
& Features     & 7.7 & 7.0 & 7.5 & 6.8 & 7.3 & 6.7 \\
& \textbf{Average} & \textbf{7.15} & \textbf{6.53} & \textbf{6.90} & \textbf{6.30} & \textbf{6.85} & \textbf{6.28} \\
\midrule
\rowcolor{gray!15} \multicolumn{8}{c}{\textbf{Backbone Model: Qwen2.5-VL-3B}} \\
\midrule
\multirow{4}{*}{Pretrained} 
& Treatment     & 6.1 & 5.3 & 6.2 & 5.1 & 5.2 & 4.4 \\
& Causes       & 4.2 & 3.7 & 6.1 & 5.0 & 5.1 & 4.2 \\
& Demographics & 6.0 & 5.2 & 6.2 & 5.3 & 6.1 & 5.2 \\
& Features     & 7.9 & 6.3 & 8.1 & 6.7 & 8.0 & 6.6 \\
& \textbf{Average} & \textbf{6.05} & \textbf{5.13} & \textbf{6.65} & \textbf{5.53} & \textbf{6.10} & \textbf{5.10} \\
\midrule
\multirow{4}{*}{\shortstack{Conversation \\ Tuned}} 
& Treatment     & 7.9 & 7.2 & 8.0 & 7.1 & 8.0 & 7.1 \\
& Causes       & 8.1 & 7.3 & 8.1 & 7.4 & 7.1 & 6.5 \\
& Demographics & 7.0 & 6.2 & 5.1 & 4.7 & 6.2 & 5.7 \\
& Features     & 8.0 & 7.1 & 8.1 & 7.2 & 7.1 & 6.4 \\
& \textbf{Average} & \textbf{7.75} & \textbf{6.95} & \textbf{7.33} & \textbf{6.60} & \textbf{7.10} & \textbf{6.43} \\
\midrule
\multirow{4}{*}{DermIQ-VLM} 
& Treatment     & 8.6 & 7.7 & 8.1 & 7.4 & 8.6 & 7.8 \\
& Causes       & 8.4 & 7.6 & 8.5 & 7.7 & 8.3 & 7.6 \\
& Demographics & 8.2 & 7.4 & 8.3 & 7.5 & 8.2 & 7.5 \\
& Features     & 8.9 & 7.8 & 8.8 & 7.9 & 8.7 & 7.8 \\
& \textbf{Average} & \textbf{8.53} & \textbf{7.63} & \textbf{8.42} & \textbf{7.63} & \textbf{8.45} & \textbf{7.68} \\
\bottomrule
\end{tabular}%
}
\caption{Evaluation of conversational responses using Accuracy, Relevance, and Completeness metrics, judged by Grok and GPT-4, for \textbf{Qwen2-VL-2B} (top) and \textbf{Qwen2.5-VL-3B} (bottom). Backbone model rows are highlighted in silver.}
\label{tab:rag_performance_merged_silver}
\end{table}

\subsection{\textbf{Performance of Conversational Quality}}

Conversation tuning led to clear improvements in conversational quality across both model backbones, as summarized in Table~\ref{tab:rag_performance_merged_silver}. While both backbones benefited, larger models showed more pronounced gains in accuracy and completeness. Integrating knowledge-graph-based Direct Preference Optimization (DPO) further amplified these improvements, enabling the final \textbf{DermIQ-VLM} to consistently deliver the most reliable and coherent outputs.

Assessments by Grok and GPT-4 remained closely aligned, suggesting robustness across evaluators. The results illustrate a clear progression: conversational fine-tuning aligns models with domain-specific dialogue patterns, and retrieval-guided preference optimization systematically elevates response quality, producing context-aware responses suitable for complex biomedical conversational tasks.

\section{Limitations and Future Work}
Despite promising results, GRPO++ faces limitations: performance depends on dataset size and quality, GPU constraints restricted potential gains, and the computationally costly multi-stage training pipeline hinders scalability. Validation has been mainly in dermatology, raising generalization concerns, and reliance on curated knowledge graphs introduces coverage and consistency issues. Future work will extend GRPO++ to structured reasoning domains, refine it for complex tasks, improve scalability with larger datasets and models, and enhance reliability through expanded knowledge bases and optimized training, broadening its applicability across clinical and reasoning-intensive fields.

\section{Conclusion}
This study tackles the challenge of building medical vision-language models (VLMs) for low-resource settings, emphasizing explainable and accurate skin disease detection. We present DermIQ-VLM, a VLM designed to emulate dermatologists’ diagnostic reasoning. Key innovations include a memory-efficient variant of Group Relative Policy Optimization (GRPO++) and multi-stage training with supervised fine-tuning, Knowledge Graph Retrieval-Augmented Generation (KG-RAG), and Direct Preference Optimization (DPO). Together, these methods significantly improve diagnostic ability, achieving a 52\% detection rate under limited data. Evaluations by benchmark LLMs confirmed gains in factual accuracy, relevance, and completeness. While results are encouraging, further progress depends on larger, higher-quality datasets and expanded domain knowledge.

\section{Acknowledgement}
We would like to acknowledge that we have utilized AI models to assist in refining our grammatical mistakes and enhancing the conversational flow.

\end{document}